\documentclass[]{jingdong}

\usepackage{amsfonts}
\usepackage{nicefrac}
\usepackage{enumitem}
\usepackage{pifont}

\usepackage{amsmath,amssymb}
\usepackage{amsthm}
\usepackage{algpseudocode}
\usepackage{colortbl}
\usepackage{wrapfig}
\usepackage{booktabs}
\usepackage{tabularx}
\usepackage{array}
\usepackage[table]{xcolor}
\usepackage[utf8]{inputenc}

\usepackage{algorithm}
\usepackage{multirow}

\microtypesetup{expansion=false}
\graphicspath{{figures/}}
\definecolor{SkillSection}{HTML}{F3F6FA}
\definecolor{SkillTotal}{HTML}{E8EEF7}
\definecolor{SkillStripe}{HTML}{FAFAFA}
\definecolor{rubricblue}{HTML}{D85C55}

\title{Learning from Online User Feedback for Shopping Agents}

\author[1,\ddagger]{Haobo Zhang}
\author[2]{Kelong Mao}
\author[2]{Sulong Xu}
\author[2]{Simiu Gu}
\author[1,\dagger]{Zhicheng Dou}

\affiliation[1]{Gaoling School of Artificial Intelligence, Renmin University of China}
\affiliation[2]{JD.COM}

\checkdata[Email]{\email{zhanghb@ruc.edu.cn};\email{dou@ruc.edu.cn};  \email{maokelong.1@jd.com}} 

\abstract{Large language model-based shopping agents are increasingly deployed in real-world e-commerce platforms, generating massive amounts of user interaction logs that provide valuable supervision for improving these agents. However, existing approaches primarily rely on offline training signals, such as user-item interactions or synthetic preference data, while largely overlooking the rich supervision contained in users' natural conversational feedback. Moreover, the available online feedback is heterogeneous, sparse, and noisy, making it difficult to transform into reliable learning signals automatically.
To address these challenges, we propose LOFA, a framework that enables shopping agents to learn directly from real online interaction logs without human annotation. 
LOFA combines reinforcement learning over verifiable purchase outcomes with feedback-aware on-policy distillation, which identifies users’ in-dialogue directives and converts them into dense token-level supervision. These complementary objectives capture both collaborative behavioral patterns and user-specific preferences.
Extensive experiments on real-world e-commerce logs demonstrate that LOFA consistently improves recommendation quality, response helpfulness, and user-satisfaction alignment over strong baselines, highlighting the effectiveness of learning shopping agents from real online user feedback.
}

\begin{document}

\maketitle

\setlength{\skip\footins}{12pt}

\begingroup
\renewcommand{\thefootnote}{}
\footnotetext{
\textsuperscript{\ensuremath{\ddagger}}This work was completed during Haobo Zhang's internship at JD.COM.

\quad
\textsuperscript{\ensuremath{\dagger}}Corresponding author.
}
\endgroup

\vspace{-2mm}

\section{Introduction}

Large language models (LLMs) have recently driven the rapid development of conversational shopping agents, enabling users to search for products, compare alternatives, and make purchasing decisions through natural language interactions. Compared with traditional recommendation systems~\cite{BERT4Rec2019, SASRec2018, BPR2012, S3Rec2020} that mainly rank items according to historical behaviors, shopping agents~\cite{LLaSA2024, amazonshopagent2025} can actively understand users' evolving intents, conduct multi-turn conversations, invoke external tools such as product search engines, and provide personalized recommendations together with natural language explanations~\cite{amazonconvshop2025, AliMe2018}. As these agents are increasingly deployed in real-world e-commerce platforms such as Amazon, they continuously accumulate massive amounts of interaction logs between users and agents. Beyond recording whether a recommendation eventually leads to a purchase, these logs also capture users' natural reactions, preference refinements, and corrective feedback throughout the conversation, providing a valuable yet largely untapped source for improving shopping agents.

\begin{figure*}[t]
\centering
\includegraphics[width=0.98\linewidth]{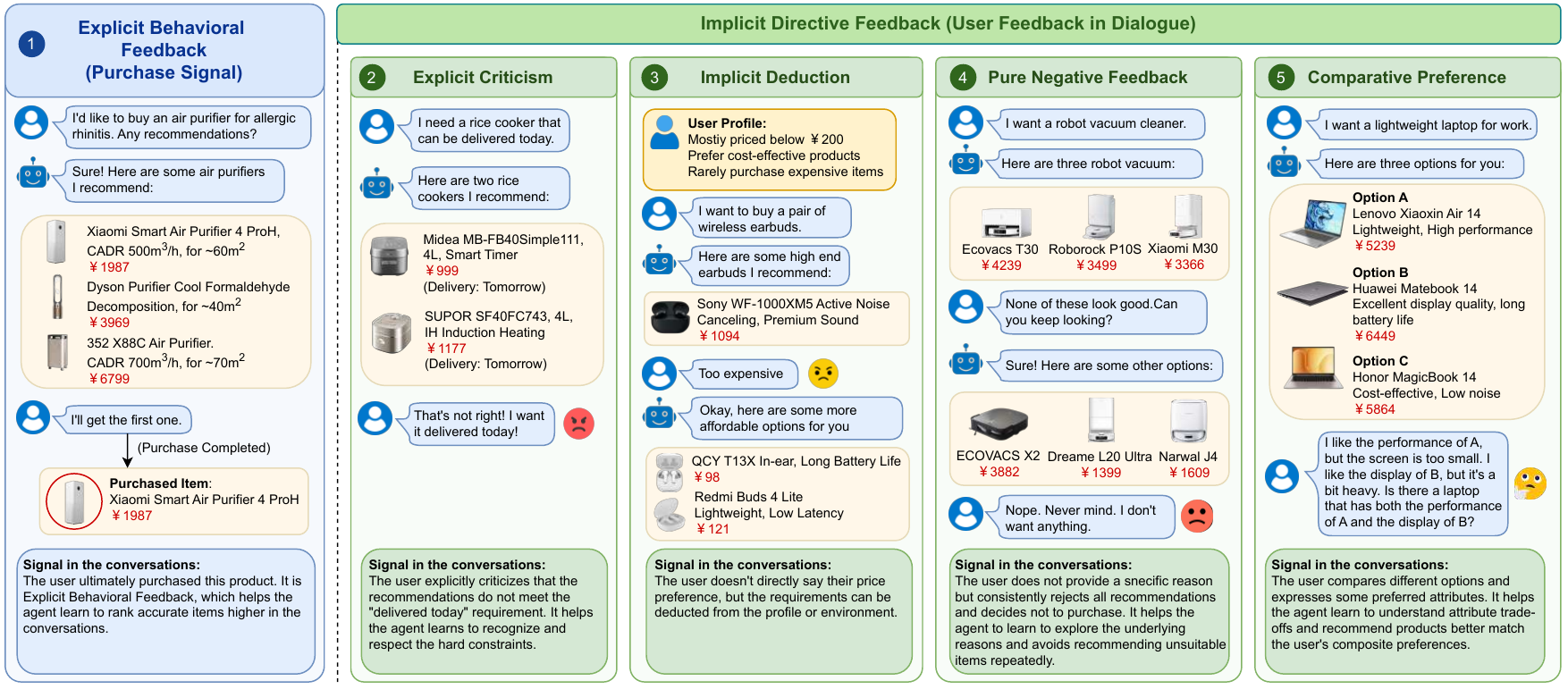}
\caption{Examples of supervision signals contained in real-world online shopping logs, which include explicit behavioral feedback (e.g., purchase outcomes) and in-dialogue directive feedback. These feedback signals reveal why users are (dis)satisfied with previous recommendations and provide fine-grained guidance for improving future recommendation decisions and conversational responses. }
\label{fig:intro}
\end{figure*}

Recent research has substantially advanced shopping agents through improvements in reasoning, planning, tool use, and personalized recommendation~\cite{R2ec2025, Reason2Recommend2025, ToolRec2024}. Meanwhile, several studies have explored more sophisticated agent architectures, such as integrating user profiles, memory modules, and multi-agent collaboration to better model user preferences and support complex recommendation scenarios~\cite{MACRS2024, MACRec2024}.
Building upon these advances in agent capabilities, existing optimization approaches mainly improve shopping agents through
supervised instruction tuning, preference optimization, or reinforcement learning over carefully curated offline datasets~\cite{Reason4Rec2025, RecThinker2026}. 
These methods primarily learn from static supervision, including user-item interaction data and synthetic preference annotations, and have achieved strong performance under offline training settings. 

Despite these advances, effectively learning from real online user feedback remains largely unexplored.
First, real-world user-agent interactions naturally generate two complementary forms of online user feedback: explicit behavioral feedback, such as clicks and purchases, and in-dialogue directive feedback, where users refine their preferences or express dissatisfaction through subsequent conversational turns.
Existing methods primarily rely on explicit behavioral feedback while largely overlooking in-dialogue directive feedback, which is also an important signal reflecting the user's intent and preference. Moreover, purchase signals are inherently sparse and delayed in the real logs, making it difficult to attribute successful outcomes to specific conversational decisions. 
Second, although in-dialogue directive feedback contains richer and more fine-grained information, it is highly heterogeneous, sparse, and noisy, making it difficult to automatically identify actionable signals and convert them into effective supervision. 
Consequently, valuable supervision embedded in real-world online interactions remains largely untapped for improving shopping agents.

Motivated by this observation, we posit that effective learning from online interaction logs requires jointly exploiting both of the feedback sources. Explicit behavioral feedback, such as purchases, provides reliable outcome-level supervision that directly reflects whether the recommendation is successful. On the other hand, in-dialogue directive feedback reveals the underlying reasons behind user satisfaction or dissatisfaction, offering fine-grained guidance for improving future responses. Rather than treating these signals independently, we view them as complementary supervision for learning shopping agents from real-world interaction logs: Explicit behavioral feedback answers whether the interaction achieves a successful outcome, whereas in-dialogue directive feedback explains why users are satisfied or dissatisfied. This insight motivates a unified framework that systematically transforms heterogeneous online feedback into effective training signals for shopping agents and enables the agent to improve itself from its daily logs.


To address these challenges, we propose LOFA, a unified framework for learning shopping agents from online user feedback without human annotation.
It systematically transforms heterogeneous online user feedback into effective supervision signals by jointly leveraging explicit behavioral feedback and in-dialogue directive feedback.
Specifically, for explicit behavioral feedback, LOFA automatically identifies interaction sessions associated with purchase outcomes from online logs and formulates them as training data. These data are then used to optimize the agent through reinforcement learning to better satisfy users' shopping needs and lead to more accurate recommendations.
For implicit conversational feedback, LOFA leverages an LLM-based feedback mining pipeline to identify available directive signals contained in subsequent user responses. We categorize these signals into four representative feedback types that reflect user dissatisfaction, preference refinement, and missing constraints. To effectively learn from such sparse and heterogeneous feedback, we further introduce an on-policy distillation framework, where subsequent user feedback is incorporated as additional guidance to construct an enhanced teacher model. The resulting teacher distributions provide dense token-level supervision, enabling the student model to learn how its previous responses should be revised according to user feedback.
By jointly learning from purchase outcomes and conversational feedback, LOFA learns not only whether recommendations succeed but also why users are satisfied or dissatisfied, leading to more accurate recommendations and responses that better align with user intent.

Our main contributions are summarized as follows:

(1)  We formulate learning from online user feedback as a new paradigm for improving shopping agents and propose \textbf{LOFA}, a unified framework that automatically transforms online user feedback extracted from real-world logs into effective supervision signals without human annotation.

(2) We identify two complementary forms of online user feedback—explicit behavioral feedback and in-dialogue directive feedback—and develop an LLM-based feedback mining framework to automatically extract actionable conversational supervision from noisy real-world interactions.

(3) We introduce a dual-feedback optimization strategy that combines reinforcement learning from purchase outcomes with on-policy distillation from conversational feedback, enabling shopping agents to learn not only whether recommendations succeed but also why users are satisfied or dissatisfied.

\section{Related Work}

\subsection{Agent-based Recommendation}

Recent advances in LLMs have inspired a new generation of agent-based RSs, which formulate recommendation as an autonomous decision-making process involving reasoning, planning, memory, and interaction with external environments. Existing studies can be broadly divided into agent-based interaction simulators and agent-based recommenders.

\textbf{Agent-based Interaction Simulation.} 
This line of work employs agents to simulate users, items, and environments to model recommendation dynamics and refine the memory of users and items~\cite{AgentCFplus2025, RecAgent2025, SimUSER2025, CSHI2025, AgentCF2023}.
They mainly focus on user behavior simulation and the modeling of user and item characteristics, which enable better simulation and characterization of real-world scenarios.

\textbf{Agent as Recommender.} 
Another line of research directly employs agents as core recommenders that interact with users and external tools to provide recommendations.
Recent studies have enhanced recommendation agents through improved reasoning, planning, and tool-use capabilities~\cite{Reason2Recommend2025, ToolRec2024, RecThinker2026, RecMind2024}. 
Other studies focus on designing specialized modules to better capture user preferences and support complex recommendation scenarios, such as user profiling~\cite{PersonaX2025}, memory mechanisms~\cite{InteRecAgent2025}, and multi-agent collaboration~\cite{MACRS2024, MACRec2024}. More recently, several works have explored optimization strategies including instruction tuning, preference optimization, and reinforcement learning to further improve recommendation quality~\cite{Reason4Rec2025, RecThinker2026}.

Despite substantial progress in agent capabilities, existing methods primarily focus on improving recommendations through architectural design and reasoning strategies. 
Most of them rely on static training datasets or explicit interactions, while paying limited attention to the rich feedback in real-world user-agent interactions. 
In contrast, our work investigates how shopping agents can automatically learn from online interaction logs by exploiting both behavioral outcomes and conversational feedback after deployment.

\subsection{Learning from Feedback for LLM Agents}

Learning from feedback has become a key paradigm for improving large language models. RLHF~\cite{RLHF1_2017, RLHF2_2017} aligns model behaviors with human preferences through reinforcement learning, while subsequent approaches such as DPO~\cite{DPO2023} and GRPO~\cite{GRPO2024} further improve the scalability and effectiveness of preference optimization. 
Another related direction explores knowledge transfer through on-policy distillation~\cite{OPD2024, KDsurvey2024, MiniLLM-KD2024}.
These methods~\cite{OPD2024, OPSD} provide dense token-level supervision from a teacher model to train the visitation distribution of the student policy through KL-based objectives. 
However, existing feedback-learning methods mainly rely on curated preference datasets, reasoning traces, or outcome-level rewards. Our work instead focuses on automatically learning from real online user feedback, jointly leveraging behavioral outcomes and in-dialogue directive feedback extracted from deployed shopping agents.

\section{Method}

We propose \textbf{LOFA}, a unified framework that automatically learns from online user feedback by jointly leveraging explicit behavioral feedback and in-dialogue directive feedback. In this section, we first introduce the task formulation and overall framework, and then present the learning strategies for the two feedback sources.

\begin{figure*}[t]
\centering
\includegraphics[width=0.97\linewidth]{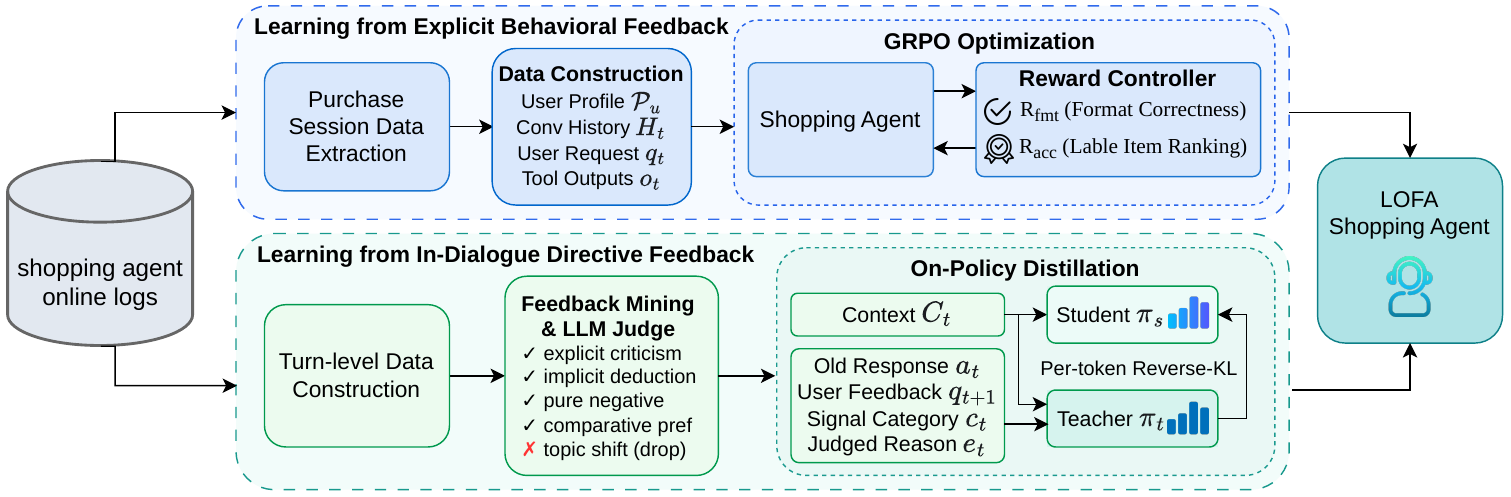}
\caption{The overall architecture of our LOFA framework. }
\label{model_graph}
\end{figure*}

\subsection{Preliminaries}

\subsubsection{Notation and Task Formulation}

We consider an online shopping agent that interacts with users through multi-turn conversations and external product search tools.

For a user $u$, we denote the user profile as $\mathcal{P}_u$, which consists of a meta profile, a short-term preference profile, and a shopping memory. An interaction session is represented as

\[
S = \{(q_1,o_1,a_1), \ldots, (q_T,o_T,a_T)\},
\]
where $q_t$ denotes the user query at turn $t$, $a_t$ denotes the agent response, and $o_t$ denotes the product search results returned by external tools. Each of the tool response $o_t$ is composed of several candidates $\{c_{1},..., c_{n}\}$.
For sessions with successful purchases, we further denote the purchased item as $i^*$.

There are two main tasks in our model, where LOFA learns from two complementary forms of supervision extracted from real-world interaction logs.

\textbf{Behavioral Feedback Learning.}
Given an interaction session $S$ and its purchase outcome $i^*$, the objective is to optimize the shopping agent to rank the purchased item higher, thereby generating recommendations that better satisfy user needs.

\textbf{Directive Feedback Learning.}
Given a dialogue turn $(q_t,o_t)$ and its session context $S$ and the subsequent user feedback $q_{t+1}$, 
the goal is to extract actionable preference signals and train the agent to generate responses $a_t'$ that better address user needs at the current turn $t$.

\subsubsection{Overview}

Figure~\ref{model_graph} illustrates the overall framework of LOFA. Starting from large-scale online interaction logs collected from deployed shopping agents, LOFA automatically extracts two complementary supervision signals: explicit behavioral feedback and in-dialogue directive feedback.

For explicit behavioral feedback, LOFA constructs session-level recommendation data from interaction sessions that result in successful purchases. The agent is then optimized through reinforcement learning with purchase-based rewards, enabling it to better identify products that match user preferences and lead to successful conversions.

For in-dialogue directive feedback, LOFA constructs turn-level training instances from multi-turn interaction sessions. We first employ an LLM-based feedback mining module to identify actionable directive signals contained in subsequent user responses. The extracted feedback is then incorporated into a feedback-aware teacher model, which provides token-level guidance through on-policy distillation and helps the agent learn how previous responses should be revised.

Since the two feedback sources are available at different granularities and require different optimization objectives, LOFA adopts a sequential training pipeline. The agent is first optimized with reinforcement learning from purchase outcomes and is subsequently refined through self-distillation from conversational feedback, enabling continuous improvement from both outcome-level and turn-level supervision.

\subsection{Learning from Explicit Behavioral Feedback}

Purchase outcomes provide reliable outcome-level supervision that directly reflects whether a shopping agent successfully satisfies user needs. Therefore, LOFA first learns from explicit behavioral feedback by optimizing the agent on interaction sessions that lead to successful purchases.

\subsubsection{Behavioral Feedback Construction}

We construct reinforcement learning data from real-world shopping sessions collected from deployed shopping agents. Specifically, we retain sessions that end with successful purchases and extract the complete interaction trajectory together with the corresponding user profile, including the meta profile, short-term preferences, and shopping memory. Given the interaction context and retrieved product candidates, the agent is trained to generate personalized responses and recommendations.


Formally, for a session $S=\{(q_1,o_1,a_1),\ldots,(q_T,o_T,a_T)\}$ with purchased item $i^*$, the agent predicts a response $a_t'$ and a ranked recommendation list $R_t=\{i_1,i_2,\ldots,i_K\}$ based on the available interaction context. The purchase outcome $i^*$ serves as a verifiable reward signal for reinforcement learning.

\subsubsection{GRPO Optimization}

We optimize the shopping agent using the GRPO algorithm. 
To ensure both recommendation accuracy and response validity and quality, we design two complementary reward functions.


\textbf{Format Reward.}
Shopping agents are required to follow a predefined response format. 
We assign a penalty to responses that violate the predefined output format, including missing \texttt{<think>} or \texttt{<answer>} tags, insufficient recommendations, or invalid product identifiers.
Specifically, given a trajectory $\tau$,  the format reward is defined as:
\begin{equation}~\label{sec:formatreward}
R_{\mathrm{fmt}}(\tau) =
\begin{cases}
0, & \text{if } \tau \text{ follows the predefined format}, \\
-1, & \text{otherwise}.
\end{cases}
\end{equation}

\textbf{Recommendation Reward.}
We further evaluate recommendation quality according to the ranking position of the label item. Specifically, given the generated recommendation list $R_t$, we compute NDCG@10 as the reward using the purchased item $i^*$ as the target item. 
Formally, the recommendation reward is defined as:
\begin{equation}~\label{sec:AccReward}
R_{\mathrm{rank}}(\tau) = \mathrm{NDCG@10}(\tau).
\end{equation}

The overall reward is computed as:
\begin{equation}
R(\tau) = 
\begin{cases} 
-1, & \text{if } R_{\mathrm{fmt}}(\tau) = -1, \\
R_{\mathrm{rank}}(\tau), & \text{otherwise}.
\end{cases}
\end{equation}

By optimizing the agent with GRPO on purchase-supervised sessions, LOFA learns recommendation behaviors that better satisfy users' shopping needs.

\subsection{Learning from In-Dialogue Directive Feedback}

While behavioral feedback provides reliable outcome-level supervision, it is inherently sparse and delayed. In contrast, users frequently express dissatisfaction, preference refinements, and additional requirements during subsequent conversational turns. Such feedback reveals not only whether a recommendation succeeds, but also why it fails. Therefore, LOFA further learns from in-dialogue directive feedback to capture fine-grained user preferences and improve response quality.

\subsubsection{Feedback Mining}

We construct directive-feedback data from multi-turn interaction sessions collected from real-world interaction logs. Unlike behavioral feedback learning, we do not require successful purchases and retain both successful and unsuccessful conversations.

Given a session $S=\{(q_1,o_1,a_1),\ldots,(q_T,o_T,a_T)\}$,
we decompose each turn into an independent training instance. For a target turn $t$, we construct the sample using the user profile, conversation history, historical tool interactions, current user request $q_t$, current tool results $o_t$, the agent response $a_t$, and the subsequent user response $q_{t+1}$. For the final turn, where no subsequent user response exists, we replace $q_{t+1}$ with a terminal outcome description indicating whether the user eventually purchased a product.

We then employ an LLM to analyze the interaction and identify the type of directive signal contained in the subsequent feedback. Specifically, we categorize feedback into five representative types:

\begin{itemize}
\item \textbf{Explicit Criticism}: the user explicitly points out violations of previously stated requirements.
\item \textbf{Implicit Deduction}: the user provides constraints that could have been inferred from profiles, context, or commonsense knowledge.
\item \textbf{Pure Negative Feedback}: the user rejects the recommendation without providing specific improvement directions.
\item \textbf{Comparative Preference}: the user reveals fine-grained preferences through comparisons among products.
\item \textbf{Topic Shift/Abandonment}: the user changes topics or abandons the shopping task.
\end{itemize}

Among these categories, the first four contain actionable supervision signals that indicate how the agent should improve its response. In contrast, Topic Shift/Abandonment provides little information about recommendation quality and is therefore discarded. The LLM further generates a concise explanation describing the potential reason behind the user's dissatisfaction or preference refinement. We denote the identified feedback category as $c_t$ and the generated explanation as $e_t$, which are subsequently used to construct feedback-aware supervision.

\subsubsection{Feedback-Aware Teacher Construction}

A key challenge in utilizing directive feedback is converting sparse natural-language reactions into effective learning signals. Instead of directly optimizing on user feedback, LOFA constructs a feedback-aware teacher that has access to privileged future information and provides corrective guidance for previous responses.

We first define the deployment-time context as $C_t=[\mathcal{P}_u;H_t;q_t;o_t]$, where $\mathcal{P}_u$ denotes the user profile, $H_t$ denotes the conversation history together with historical tool interactions, $q_t$ denotes the current user request, and $o_t$ denotes the current tool outputs. 
The student model only observes deployment-time information: $X_t^{S}=C_t$.

The teacher is constructed by augmenting the deployment-time context with privileged feedback information: $X_t^{T} = [C_t; a_t; q_{t+1}; c_t; e_t]$, where the additional inputs include the previous agent response $a_t$, subsequent user feedback $q_{t+1}$, the directive feedback category $c_t$, and the generated explanation $e_t$. We further prepend a teacher instruction that encourages the model to analyze how the previous response should be improved under the observed feedback.

Unlike conventional knowledge distillation that relies on an external teacher, LOFA constructs a feedback-aware teacher from the same underlying model using privileged future feedback. This design enables the teacher to generate token-level corrective signals that explicitly reflect user feedback and preference refinement.

\subsubsection{On-Policy Distillation}

After constructing the feedback-aware teacher, we perform on-policy distillation to transfer the corrective knowledge to the student model.
Let $\hat a_t=\{\hat a_1,\ldots,\hat a_N\}$ denote a response generated by the student policy. We define the student and teacher token distributions at generation step $j$ as \[ p_s^{(j)} = p_s(\cdot \mid X_t^{S},\hat a_{<j}), \] \[ p_t^{(j)} = p_t(\cdot \mid X_t^{T},\hat a_{<j}), \] respectively. 
Rather than generating an alternative response, the teacher evaluates the student's trajectory under the privileged feedback context and provides token-level guidance.

We optimize the student using reverse KL divergence:
\[ 
    \mathcal L_{\text{OPD}} = \mathbb E_{\hat a_t\sim p_s} \left[ D(p_s \,\|\, p_t) \right], 
\] 
where the sequence-level divergence is computed as the average token-level divergence: 
\[ 
    D(p_s \,\|\, p_t) = \frac{1}{N} \sum_{j=1}^{N} D_{\mathrm{KL}} \left( p_s^{(j)} \,\|\, p_t^{(j)} \right). 
\] 
It enables the teacher to provide dense token-level supervision throughout the generation process. 
In this way, sparse conversational feedback is transformed into fine-grained optimization signals, enabling the student to generate responses that better align with user preferences.




\subsection{Training Pipeline}

The two feedback sources considered in LOFA exhibit fundamentally different characteristics. Explicit behavioral feedback is available at the session level and requires successful purchase outcomes, whereas directive feedback is extracted at the turn level and can be obtained from both successful and unsuccessful interactions. Consequently, the two learning objectives operate on different data granularities and supervision signals, making direct joint optimization less suitable.

Therefore, LOFA adopts a sequential training pipeline. We first optimize the shopping agent with GRPO using purchase signals, enabling the model to learn recommendation behaviors that lead to successful purchases. The resulting model is then further refined through on-policy distillation using directive feedback, allowing it to better capture user preferences and improve response quality. Since both forms of supervision can be automatically extracted from real-world interaction logs without human annotation, LOFA naturally supports periodic model updates as new online user feedback becomes available.

\section{Experiments}

\begin{table}[t]
\small
\centering
\caption{Statistics of the dataset for behavioral feedback learning.}
\setlength{\tabcolsep}{4.5mm}{
\begin{tabular}{lrrrrr}
\toprule
Datasets & \#Users & \#candidates & \#Purchased i & \#Sessions & \#Avg turns \\
\midrule
JD-Search & 6,704 & 125,694 & 6,685 & 7,001 & 2.07 \\
\quad$\hookrightarrow$ train & 6,059 & 114,086 & 6,039 & 6,303 & 2.07 \\
\quad$\hookrightarrow$ test & 692 & 14,393 & 693 & 698 & 2.09 \\
\bottomrule
\end{tabular}
}
\label{tab:rl_data}
\end{table}

\begin{table}[t]
\small
\centering
\caption{Statistics of the dataset for directive feedback learning. \#EC, \#ID, \#PNF, and \#CP denote the numbers of Explicit Criticism, Implicit Deduction, Pure Negative Feedback, and Comparative Preference instances, respectively.}
\setlength{\tabcolsep}{4.0mm}{
\begin{tabular}{lrrrrrrr}
\toprule
Datasets & \#Users & \#Sessions & \#Samples & \#EC & \#ID & \#PNF & \#CP \\
\midrule
JD-conv & 6,019 & 6,121 & 8,249 & 673 & 5,413 & 1,987 & 176 \\
\quad$\hookrightarrow$ train & 5,433 & 5,511 & 7,437 & 608 & 4,888 & 1,782 & 159 \\
\quad$\hookrightarrow$ test & 610 & 610 & 812 & 65 & 525 & 205 & 17 \\
\bottomrule
\end{tabular}
}
\label{tab:opd_data}
\end{table}

\subsection{Experimental Setup}

\begin{table*}[!t]
\footnotesize
\centering
\renewcommand{\arraystretch}{1.08}
\caption{The results of NDCG, Recall, MAP@1, 10, 20, and overall success rate on two tasks. The best results are shown in bold.}
\setlength{\tabcolsep}{4.8pt}{
\begin{tabular}{lcccccccccc}
\toprule
\multicolumn{1}{c}{\multirow{2}{*}{Method}} & \multicolumn{9}{c}{JD-Search} & JD-conv \\
\cmidrule(lr){2-10}\cmidrule(lr){11-11}
 & N@1 & N@5 & N@10 & R@1 & R@5 & R@10 & MAP@1 & MAP@5 & MAP@10 & SR \\
 \midrule
Qwen3-8B (NoThink) & 0.2479 & 0.3948 & 0.4039 & 0.2479 & 0.5287 & 0.5573 & 0.2479 & 0.3504 & 0.3540 & 0.3855 \\
Qwen3-8B & 0.2679 & 0.4390 & 0.4437 & 0.2679 & 0.5959 & 0.6103 & 0.2679 & 0.3871 & 0.3890 & 0.5468 \\
Qwen3-8B-reflect & 0.2765 & 0.4457 & 0.4515 & 0.2765 & 0.6017 & 0.6161 & 0.2765 & 0.3924 & 0.3945 & 0.5542 \\
Qwen3-8B-SFT & 0.2865 & 0.4358 & 0.4456 & 0.2865 & 0.5731 & 0.6032 & 0.2865 & 0.3904 & 0.3945 & 0.4064 \\
Qwen3-8B-RL & 0.4112 & 0.6168 & 0.6389 & 0.4112 & 0.7894 & 0.8567 & 0.4112 & 0.5590 & 0.5683 & 0.5677 \\
Qwen3-8B-OPD & 0.2779 & 0.4341 & 0.4417 & 0.2779 & 0.5731 & 0.5959 & 0.2779 & 0.3879 & 0.3912 & 0.6010 \\
Qwen3-8B-OPD$\rightarrow$RL & 0.4197 & 0.6107 & 0.6386 & 0.4197 & 0.7736 & 0.8595 & 0.4197 & 0.5562 & 0.5677 & 0.5924 \\
Qwen3-8B-RL$\rightarrow$OPD & {\scriptsize \textbf{0.4212}} & \textbf{0.6189} & \textbf{0.6512} & \textbf{0.4212} & \textbf{0.7923} & \textbf{0.8911} & \textbf{0.4212} & \textbf{0.5612} & \textbf{0.5747} & \textbf{0.6022} \\
\bottomrule
\end{tabular}
}
\label{tab:main_results}
\end{table*}

\subsubsection{Datasets}




We construct two datasets from real-world interaction logs collected from Jingyan, a deployed shopping agent on JD.com. To protect user privacy, all data are anonymized and randomly sampled from a fixed time period.
The behavioral-feedback dataset consists of interaction sessions ending in successful purchases, each containing complete conversations, tool interactions, retrieved products, and the purchased item. The directive-feedback dataset is constructed by decomposing multi-turn conversations into turn-level training instances, each annotated with a directive-feedback category and rationale using the feedback mining pipeline. Tables~\ref{tab:rl_data} and \ref{tab:opd_data} summarize the dataset statistics.

\subsubsection{Compared Methods}

We adopt Qwen3-8B~\cite{Qwen3} as the backbone shopping agent and evaluate the effectiveness of different learning strategies:

\begin{itemize}
\item \textbf{Base}: the original Qwen3-8B shopping agent without additional optimization.
\item \textbf{NoThink}: The same backbone model with the reasoning mode disabled during inference.
\item \textbf{+SFT}: A supervised fine-tuning baseline trained on high-quality trajectories selected from online logs. Due to the absence of reasoning traces in the logs, the model is trained solely on responses without explicit thinking processes.
\item \textbf{Self-Reflection}: an inference-time self-refinement baseline that revises responses by asking the agent to reflect on its output without parameter updates.
\item \textbf{+RL}: the model optimized only with behavioral feedback through GRPO.
\item \textbf{+OPD}: the model optimized only with directive feedback through on-policy distillation.
\item \textbf{+RL$\rightarrow$OPD}: the full LOFA framework that first performs GRPO and subsequently applies OPD.
\item \textbf{+OPD$\rightarrow$RL}: a reversed training order variant.
\end{itemize}


\subsubsection{Evaluation Metrics}

For behavioral feedback learning, we evaluate recommendation quality using NDCG@K, Recall@K, and MAP@K with $K\in\{1,5,10\}$. The purchased item is treated as the target item, and ranking metrics are computed according to its position in the final recommendation list generated by the agent. 
Note that the length of the ranking list is not always 10 but depends on the decision of the agent.

For directive feedback learning, we evaluate whether the revised response successfully addresses the issue revealed by the subsequent user question. Specifically, we prompt an external LLM evaluator with the interaction context, the original response, the user feedback, and the revised response, and compute the Success Rate (SR), which measures the proportion of responses that successfully resolve the identified issue.

\subsubsection{Implementation Details}

We use Qwen3-8B as the backbone model. During training, we set the sampling temperature to 1.0 and top-p to 0.95 to encourage exploration, while inference uses temperature 0.6 and top-p as 0.95.
For both GRPO and OPD training, we adopt full-parameter fine-tuning with a learning rate of $1\times10^{-6}$ and a batch size of 32. Models are trained for 1-2 epochs. In GRPO, the group size is set to $G=5$. All results are averaged over multiple runs at the evaluation stage.
For directive-feedback mining and response evaluation, we use DeepSeek-V3.2~\cite{DeepSeek-V3.2} as the judge model.

\subsection{Main Results}

Table~\ref{tab:main_results} reports the overall performance of different training strategies on both recommendation quality and directive-feedback resolution.
Several observations can be drawn from the results.

(1) Learning from explicit behavioral feedback substantially improves recommendation quality. Compared with the Base model, \textbf{+RL} consistently improves NDCG, Recall, and MAP, demonstrating that purchase outcomes provide reliable outcome-level supervision for optimizing shopping agents.

(2) Learning from in-dialogue directive feedback substantially improves the agent's ability to address user concerns and preference refinements. Compared with the Base model, \textbf{+OPD} achieves a much higher Success Rate, showing that LOFA effectively transforms directive feedback into actionable token-level supervision.



(3) The two forms of online user feedback are complementary. \textbf{+RL} mainly improves recommendation quality, whereas \textbf{+OPD} primarily enhances response quality. Combining both objectives achieves the best overall performance. \textbf{RL$\rightarrow$OPD} consistently outperforms \textbf{OPD$\rightarrow$RL}, suggesting that learning reliable outcome-level supervision before fine-grained directive supervision provides a more effective optimization strategy.

(4) Both \textbf{RL$\rightarrow$OPD} and \textbf{OPD$\rightarrow$RL} substantially outperform the Base model and the inference-time \textbf{Self-Reflection} baseline. This result highlights that learning from online user feedback through model optimization is more effective than relying solely on inference-time self-correction, highlighting the value of feedback-driven learning.


(6) Both \textbf{Qwen3-8B (No-Thinking)} and \textbf{+SFT} perform substantially worse than all feedback-learning variants. The inferior performance of No-Thinking highlights the importance of explicit reasoning in shopping agents. Although +SFT outperforms No-Thinking, it remains consistently inferior to RL- and OPD-based optimization. We attribute this gap to the loss of reasoning traces during SFT, which limits the model's ability to perform multi-step reasoning. In contrast, RL and OPD directly optimize the model through online user feedback while preserving its reasoning capability, leading to consistently better performance.


\begin{table*}[t]
\footnotesize
\centering
\renewcommand{\arraystretch}{1.08}
\caption{Success Rate results on four types of data and the overall data.}
\setlength{\tabcolsep}{4.4pt}{
\begin{tabular}{lccccc}
\toprule
Method & Explicit Criticism & Implicit Deduction & Pure Negative Feedback & Comparative Preference & Overall \\
\midrule
Qwen3-8B (NoThink) & 0.3385 & 0.3219 & 0.5805 & 0.1765 & 0.3855 \\
Qwen3-8B & 0.5231 & 0.4610 & 0.7805 & 0.3529 & 0.5443 \\
Qwen3-8B-reflect & 0.5385 & 0.4724 & 0.7854 & 0.3529 & 0.5542 \\
Qwen3-8B-SFT & 0.4154 & 0.3333 & 0.6049 & 0.2353 & 0.4064 \\
Qwen3-8B-RL & 0.4462 & 0.4952 & 0.7951 & 0.5294 & 0.5677 \\
Qwen3-8B-OPD & 0.5231 & 0.5314 & 0.8049 & \textbf{0.5882} & 0.6010 \\
Qwen3-8B-OPD$\rightarrow$RL & \textbf{0.5846} & 0.5105 & \textbf{0.8098} & 0.5294 & 0.5924 \\
Qwen3-8B-RL$\rightarrow$OPD & 0.5692 & \textbf{0.5333} & 0.8000 & 0.4706 & \textbf{0.6022} \\
\bottomrule
\end{tabular}
}
\label{tab:type_results}
\end{table*}

\begin{table*}[t]
\footnotesize
\centering
\renewcommand{\arraystretch}{1.08}
\caption{Performance (NDCG, Recall, MAP@1, 10, 20 and overall success rate) of ablation models with different RL rewards.}
\setlength{\tabcolsep}{5.1pt}{
\begin{tabular}{lcccccccccc}
\toprule
Method & N@1 & N@5 & N@10 & R@1 & R@5 & R@10 & MAP@1 & MAP@5 & MAP@10 & SR \\
 \midrule
LOFA-RL & 0.4112 & 0.6168 & 0.6389 & 0.4112 & 0.7894 & 0.8567 & 0.4112 & 0.5590 & 0.5683 & 0.5677 \\
\quad \textit{w/o.} Rec reward & 0.2751 & 0.4369 & 0.4393 & 0.2751 & 0.5845 & 0.5917 & 0.2751 & 0.3881 & 0.3890 & 0.5665 \\
\quad \textit{w/o.} Fmt reward & 0.3911 & 0.5904 & 0.6276 & 0.3911 & 0.7650 & 0.8481 & 0.3911 & 0.5323 & 0.5579 & 0.5567 \\
 \midrule
LOFA-RL$\rightarrow$OPD & \textbf{0.4212} & \textbf{0.6189} & \textbf{0.6512} & \textbf{0.4212} & \textbf{0.7923} & \textbf{0.8911} & \textbf{0.4212} & \textbf{0.5612} & \textbf{0.5747} & \textbf{0.6022} \\
\quad \textit{w/o.} Rec reward & 0.2808 & 0.4469 & 0.4552 & 0.2808 & 0.5974 & 0.6232 & 0.2808 & 0.3969 & 0.4003 & 0.5751 \\
\quad \textit{w/o.} Fmt reward & 0.4040 & 0.5912 & 0.6463 & 0.4040 & 0.7708 & 0.8854 & 0.4040 & 0.5446 & 0.5693 & 0.5985 \\
\midrule
LOFA-OPD$\rightarrow$RL & 0.4197 & 0.6107 & 0.6386 & 0.4197 & 0.7736 & 0.8595 & 0.4197 & 0.5562 & 0.5677 & 0.5924 \\
\quad \textit{w/o.} Rec reward & 0.2665 & 0.4311 & 0.4381 & 0.2665 & 0.5860 & 0.6074 & 0.2665 & 0.3799 & 0.3828 & 0.5653 \\
\quad \textit{w/o.} Fmt reward & 0.4097 & 0.5795 & 0.6341 & 0.4097 & 0.7679 & 0.8510 & 0.4097 & 0.5392 & 0.5537 & 0.5813 \\
\bottomrule
\end{tabular}
}
\label{tab:rl_ablation}
\end{table*}

\subsection{Performance on Different Directive Feedback Types}

Table~\ref{tab:type_results} compares the Success Rate across different directive-feedback categories. OPD-based methods consistently outperform all baselines, showing that LOFA can effectively learn from diverse forms of in-dialogue feedback. The largest gains are achieved on \textit{Explicit Criticism} and \textit{Implicit Deduction}, as these categories provide relatively clear signals about recommendation errors or missing user preferences. Gains on \textit{Comparative Preference} are slightly smaller because such feedback requires modeling more fine-grained trade-offs among products. \textit{Pure Negative Feedback} is the most challenging category, since users reject the recommendation without indicating how it should be improved. Nevertheless, LOFA still delivers consistent improvements, suggesting that the feedback-aware teacher can extract useful corrective supervision even from ambiguous feedback.

Consistent with the observations in Table~\ref{tab:main_results}, NoThink and +SFT remain noticeably inferior to OPD-based methods across all feedback categories, further confirming the importance of explicit reasoning and feedback-driven optimization.


\begin{table*}[t]
\footnotesize
\centering
\renewcommand{\arraystretch}{1.08}
\caption{Results on four data types and overall data of ablation models with different training data.}
\setlength{\tabcolsep}{4.0pt}{
\begin{tabular}{lccccccccccc}
\toprule
\multicolumn{1}{c}{\multirow{2}{*}{Method}} & \multicolumn{6}{c}{JD-Search} & \multicolumn{5}{c}{JD-conv} \\
\cmidrule(lr){2-7}\cmidrule(lr){8-12}
 & N@1 & N@10 &R@1 & R@10 &MAP@1 & MAP@10 & Overall & EC & ID & PNF & CP \\
 \midrule
LOFA-RL$\rightarrow$OPD & \textbf{0.4212} & \textbf{0.6512} & \textbf{0.4212} & \textbf{0.8911} & \textbf{0.4212} & \textbf{0.5747} & \textbf{0.6022} & 0.5692 & \textbf{0.5333} & 0.8000 & \textbf{0.4706} \\
\quad \textit{w/o.} EC data & 0.4112 & 0.6318 & 0.4112 & 0.8553 & 0.4112 & 0.5599 & 0.5961 & 0.5231 & 0.5295 & 0.8049 & 0.4118 \\
\quad \textit{w/o.} ID data & 0.3983 & 0.6277 & 0.3983 & 0.8610 & 0.3983 & 0.5530 & 0.5653 & 0.5692 & 0.4705 & \textbf{0.8195} & 0.4118 \\
\quad \textit{w/o.} PNF data & 0.4212 & 0.6445 & 0.4212 & 0.8811 & \textbf{0.4212} & 0.5687 & 0.5800 & \textbf{0.6000} & 0.5219 & 0.7805 & \textbf{0.4706} \\
\quad \textit{w/o.} CP data & 0.4068 & 0.6266 & 0.4068 & 0.8438 & 0.4068 & 0.5567 & 0.5911 & 0.5692 & 0.5200 & 0.8000 & 0.3529 \\
\bottomrule
\end{tabular}
}
\label{tab:opd_ablation}
\end{table*}

\begin{figure*}[!t]
\centering
\includegraphics[width=0.97\linewidth]{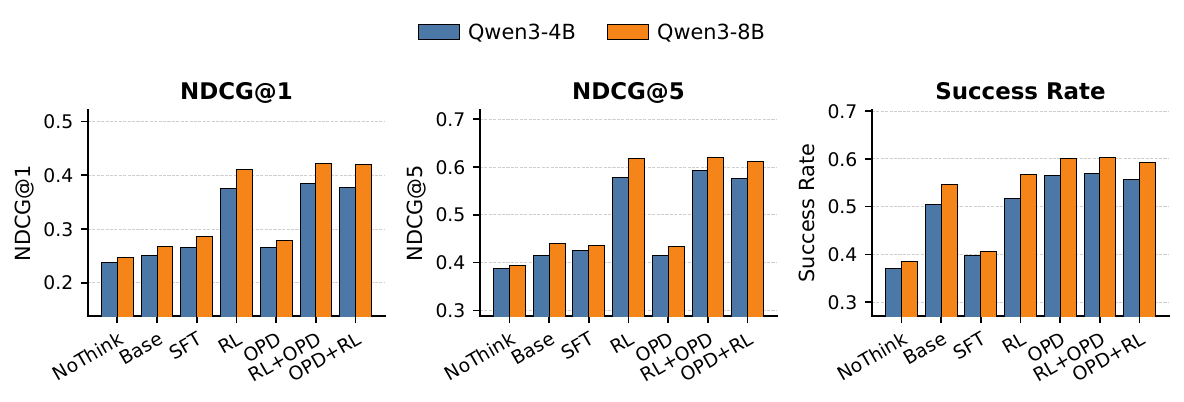}
\caption{NDCG@1 and Success Rate results of LOFA with different backbone scales on two tasks. }
\label{fig:backbone}
\end{figure*}

\subsection{Ablation Study on Behavioral Feedback Learning}

To evaluate the contribution of different reward components in GRPO, we conduct ablation studies by removing each reward: (1) \textbf{w/o. Rec Reward}: We remove the recommendation reward in Equation~(\ref{sec:AccReward}) during the RL stage.
(2) \textbf{w/o. Fmt Reward}: We remove the format reward in Equation~(\ref{sec:formatreward}) during the RL stage. Results are reported in Table~\ref{tab:rl_ablation}.
We can observe the following findings: 
Removing the recommendation accuracy reward leads to the largest performance degradation across all ranking metrics. Since this reward directly optimizes the ranking position of purchased items, its absence significantly weakens the model's ability to identify products that align with user preferences. The result confirms that recommendation accuracy is the primary learning signal for behavioral-feedback optimization.
Removing the format reward causes a relatively smaller decline. 
Although the format reward does not directly optimize recommendation quality, it encourages the model to produce valid reasoning structures and recommendation outputs, which improves training stability and reduces invalid generations.
The influence on Success Rate is relatively small, indicating that behavioral feedback learning mainly improves recommendation quality rather than conversational preference alignment.

\subsection{Ablation Study on Directive Feedback Learning}

To investigate the contribution of different directive-feedback categories, we perform ablation studies by removing one directive-feedback category at a time from the OPD training data.
Table~\ref{tab:opd_ablation} reports the results.

From the results we can find: Removing any directive-feedback category consistently reduces Success Rate, confirming that all four categories provide useful supervision. Among all categories, removing \textit{Implicit Deduction} causes the largest degradation, suggesting that latent preference constraints provide the most informative corrective signals. Removing \textit{Pure Negative Feedback} also noticeably degrades performance, indicating that even weak rejection signals contribute useful supervision. Recommendation ranking remains largely unchanged across all variants, consistent with the objective of OPD, which focuses on response quality rather than recommendation ranking.


\subsection{Generalization Across Backbone Scales}

To evaluate the generality of LOFA across different backbone scales, we additionally conduct experiments on Qwen3-4B besides the default Qwen3-8B.
For each backbone, we evaluate the same set of training strategies, including \textbf{Base}, \textbf{+RL}, \textbf{+OPD}, \textbf{+RL$\rightarrow$OPD}, and \textbf{+OPD$\rightarrow$RL}. Figure~\ref{fig:backbone} summarizes the comparison on representative recommendation and dialogue evaluation metrics.

Several observations can be made. First, the proposed LOFA framework consistently improves both recommendation quality and directive-feedback resolution on the 4B and 8B backbones, demonstrating that the effectiveness of learning from online feedback is not restricted to a particular model scale. Second, although the 8B backbone consistently achieves higher absolute performance than the 4B model, both backbones exhibit similar improvement trends after behavioral-feedback learning and directive-feedback learning. This suggests that the two forms of online supervision are complementary and remain effective across different model capacities.
These results demonstrate that the proposed framework is backbone-agnostic and can be applied to shopping agents with different parameter scales.


\begin{table*}[t]
\centering
\small
\caption{Case study on the two LOFA tasks. In \textbf{Case 1}, the original reply self-contradicts (title ``today'' vs. items ``tomorrow''), the base model silently drops the timing entirely, and our model explicitly resolves the same-day constraint. In \textbf{Case 2}, the base model narrows ``gift'' into homogeneous skincare and never recalls the target, while our model ranks the user's actual purchase first.}
\setlength{\tabcolsep}{6pt}
\renewcommand{\arraystretch}{1.2}
\begin{tabularx}{\textwidth}{@{}l >{\raggedright\arraybackslash}X @{}}
\toprule
\multicolumn{2}{@{}l}{\textbf{Case 1: Directive-Feedback Repair --- Honoring a Hard Delivery Constraint}} \\
\cmidrule(lr){1-2}
\textit{Dialogue Context} & The user is choosing a men's outfit set and states a hard delivery-time constraint. \\
\textit{User Input} & \textit{``Only ones that can be delivered \textbf{today}.''} \\
\textit{User Feedback} & No reply and no purchase in the next turn --- a complete rejection of the recommendation. \\
\cmidrule(lr){1-2}
\textit{Original Reply} &
\textbf{``Here are men's outfit sets that can be delivered today, all JD self-operated or same-day eligible:''}\newline
1. NASA MARVEL jacket set [...], \emph{pay before 23:10, expected to arrive \textbf{tomorrow}}.\newline
2. PLAYBOY hoodie set [...], \emph{pay before 18:00, expected to arrive \textbf{tomorrow}}.\newline
(Items 3--4: same pattern, all marked \emph{arrive tomorrow}.)\newline
\textcolor{red}{$\Rightarrow$``All of the above support pay-today, \textbf{arrive-tomorrow}}\,[...]''
\\
\cmidrule(lr){1-2}
\textit{Base (Qwen3-8B)} &
\textbf{``Recommended outfit sets that can be delivered today:''}\newline
1. PLAYBOY hoodie set [...], RMB 180 after coupon.\newline
2. Warrior fleece set [...], RMB 89.3, great value.\newline
3. JEEP SPIRIT set [...], RMB 138.\newline
\emph{Selection tips:} for value pick Warrior; for brand pick PLAYBOY / JEEP; check the size chart [...]. Want me to narrow it down further?\newline
\textcolor{red}{$\Rightarrow$ The header claims ``today'', but the body never mentions delivery again. (judge\_resolved = False)} \\
\cmidrule(lr){1-2}
\textit{LOFA-RL$\rightarrow$OPD} &
\textbf{``Your `outfit set' must arrive today, so I have filtered to currently in-stock items,''} ranked by value:\newline
1. American-casual set: NASA MARVEL [...]\quad 2. Versatile classic: PLAYBOY [...]\quad 3. Warm base layer: JEEP SPIRIT [...]\newline
\textbf{Note:} all screened via JD self-operated / flagship channels, supporting \textbf{30-min express delivery (JD self-operated only)}; prefer the 3 items marked \textbf{``JD own warehouse''} to avoid third-party delays. Shall I help confirm your size?\newline
\textcolor{green!55!black}{$\Rightarrow$ Directly answers the same-day constraint. (judge\_resolved = True)} \\
\midrule
\multicolumn{2}{@{}l}{\textbf{Case 2: Recommendation Ranking --- Reading the ``Gift'' Intent}} \\
\cmidrule(lr){1-2}
\textit{Dialogue Context} & \textit{``Send a gift''} $\rightarrow$ agent asks who it is for $\rightarrow$ user: \textit{``a birthday gift for a friend''}. \\
\textit{Current Query} & \textit{``A birthday gift for a friend''} \quad (candidate set: 56 items) \\
\textit{Ground Truth} & Preserved-rose flower-box ornament --- a ceremonial gift for a female friend [...]. \\
\cmidrule(lr){1-2}
\textit{Base (Qwen3-8B)} &
Ranking: 1. Est\'ee Lauder Advanced Night Repair (skincare) \enspace 2. Est\'ee Lauder serum (skincare) \enspace 3. L'Or\'eal black serum (skincare) \enspace 4. SKIN1004 centella serum (skincare) \enspace 5. panda woven keychain [...]\newline
\textcolor{red}{$\Rightarrow$ ``Gift'' collapses into ``skincare''; the flower-box is \textbf{never recalled}. (rank = None)} \\
\cmidrule(lr){1-2}
\textit{LOFA-RL$\rightarrow$OPD} &
Ranking: \textbf{1. Preserved-rose flower box} \emph{(user's actual purchase)} \enspace 2. 3D crystal-ball night-light \enspace 3. James basketball figurine \enspace 4. Est\'ee Lauder serum \enspace 5. electronic pet alarm-clock ornament [...]\newline
\textcolor{green!55!black}{$\Rightarrow$ Accurately understant the user's preference on giftable ornaments and ranks the purchase \textbf{first}. (rank = 1)} \\
\bottomrule
\end{tabularx}

\label{tab:case_study}
\end{table*}

\subsection{Case Study}

Table~\ref{tab:case_study} presents two representative cases illustrating how LOFA benefits from the two forms of online user feedback.
The first case corresponds to \textit{Pure Negative Feedback}. The user explicitly requires products that can be delivered on the same day. The original response recommends some products but the products are delivered the next day. Although the Base model changes the recommendation list, it still fails to explicitly satisfy the user's temporal constraint, resulting in an unresolved response. In contrast, LOFA correctly identifies the delivery requirement as the user's primary concern and proactively recommends products supporting same-day delivery while explaining how to avoid shipping delays. Consequently, the revised response successfully addresses the user's dissatisfaction.
The second case illustrates the effectiveness of behavioral-feedback learning. Compared with the Base model, LOFA ranks the purchased product higher and produces recommendations that better align with the user's purchasing preference. This demonstrates that reinforcement learning from purchase outcomes effectively improves recommendation quality by aligning recommendations with users' eventual purchase decisions.
These examples demonstrate that the two learning objectives are complementary. Behavioral feedback mainly improves recommendation quality, while directive feedback enables the agent to better understand user preferences and generate responses that directly resolve user concerns.

\section{Conclusion}
We presented LOFA, a framework for learning shopping agents from online user feedback without human annotation. By combining reinforcement learning over verifiable purchase outcomes with feedback-aware on-policy distillation of in-dialogue directives, LOFA improves both recommendation quality and user satisfaction. Experiments on real-world e-commerce logs demonstrate consistent gains and highlight LOFA’s practical value in enabling shopping agents to continuously learn from online interactions, forming a scalable data flywheel for industrial deployment in the future.

\bibliographystyle{unsrtnat}
\bibliography{neurips_2026}



\newpage

\end{document}